\documentclass[conference]{IEEEtran}
\usepackage{times}

\usepackage[numbers]{natbib}
\usepackage{multicol}

\usepackage{xcolor}

\newcommand{\myparagraph}[1]{\vspace{0.05in}\noindent\textbf{#1}}
\definecolor{MyDarkBlue}{rgb}{0,0.08,1}
\definecolor{MyDarkGreen}{rgb}{0.02,0.6,0.02}
\definecolor{MyDarkRed}{rgb}{0.8,0.02,0.02}
\definecolor{MyDarkOrange}{rgb}{0.40,0.2,0.02}
\definecolor{MyPurple}{rgb}{111,0,255}
\definecolor{MyRed}{rgb}{1.0,0.0,0.0}
\definecolor{MyGold}{rgb}{0.75,0.6,0.12}
\definecolor{MyDarkgray}{rgb}{0.66, 0.66, 0.66}

\newcommand{\fig}[1]{Fig.~\ref{#1}}

\newcommand{\tab}[1]{Table~\ref{#1}}

\usepackage{graphics} 
\usepackage{epsfig} 
\usepackage{mathptmx} 
\usepackage{times} 
\usepackage{amsmath} 
\usepackage{amssymb}  
\usepackage{color}
\usepackage{gensymb}
\usepackage{float}
\usepackage{amsmath}
\usepackage[pagebackref=true,breaklinks=true,letterpaper=true,colorlinks,bookmarks=false]{hyperref}

\begin{document}

\title{\LARGE \bf
GelSight Wedge: Measuring High-Resolution 3D Contact Geometry with a Compact Robot Finger
}


\author{
    \authorblockN{Shaoxiong Wang, Yu She, Branden Romero, Edward Adelson}
        \authorblockA{
     Massachusetts Institute of Technology\\
    {\tt\small <wang\_sx, yushe, brromero>@mit.edu, adelson@csail.mit.edu}} \href{http://gelsight.csail.mit.edu/wedge/}{http://gelsight.csail.mit.edu/wedge/}
\thanks{Toyota Research Institute (TRI), and the Office of Naval Research (ONR) [N00014-18-1-2815] provided funds to support this work.}
\thanks{* Authors with equal contribution.}%
}

\maketitle


\begin{abstract}

Vision-based tactile sensors have the potential to provide important contact geometry to localize the objective with visual occlusion.
However, it is challenging to measure high-resolution 3D contact geometry for a compact robot finger, to simultaneously meet optical and mechanical constraints.
In this work, we present the GelSight Wedge sensor, which is optimized to have a compact shape for robot fingers, while achieving high-resolution 3D reconstruction. 
We evaluate the 3D reconstruction under different lighting configurations, and extend the method from 3 lights to 1 or 2 lights.
We demonstrate the flexibility of the design by shrinking the sensor to the size of a human finger for fine manipulation tasks.
We also show the effectiveness and potential of the reconstructed 3D geometry for pose tracking in the 3D space.

\end{abstract}


\section{Introduction}
\label{sec:intro}

During manipulation, robots inevitably occlude the objective from vision, for example, when buttoning a shirt, or tying a shoelace. The occlusion makes it challenging for control policies with vision alone. In contrast, humans also use touch for daily manipulation tasks.

We aim to design a compact robot finger with tactile sensing to provide high-resolution 3D contact geometry, which can be used to localize~\cite{li2014localization} and estimate the pose of the in-held object with visual occlusion~\cite{izatt2017tracking, bauza2019tactile}. We leverage GelSight~\cite{yuan2017gelsight} technology in this work for its high-resolution spatial signals and 3D reconstruction.

Previous GelSight sensors~\cite{dong2017improved} provide high-resolution 3D contact geometry. However it is too bulky for robot fingers. GelSlim~\cite{donlon2018gelslim} optimizes the sensor shape, making it more compact for robotic tasks. But it is not optically designed for high-resolution 3D reconstruction. 

The major challenge is to meet optical requirements for photometric stereo, with mechanical constraints for a compact sensor. Optically, it requires to estimate gradients accurately from color signals, in different directions. Mechanically, it requires a thin and slim shape like human fingers, and having the sense of the tip to pick up small objects on the table.

\begin{figure}[t]
	\centering
	\includegraphics[width= \linewidth]{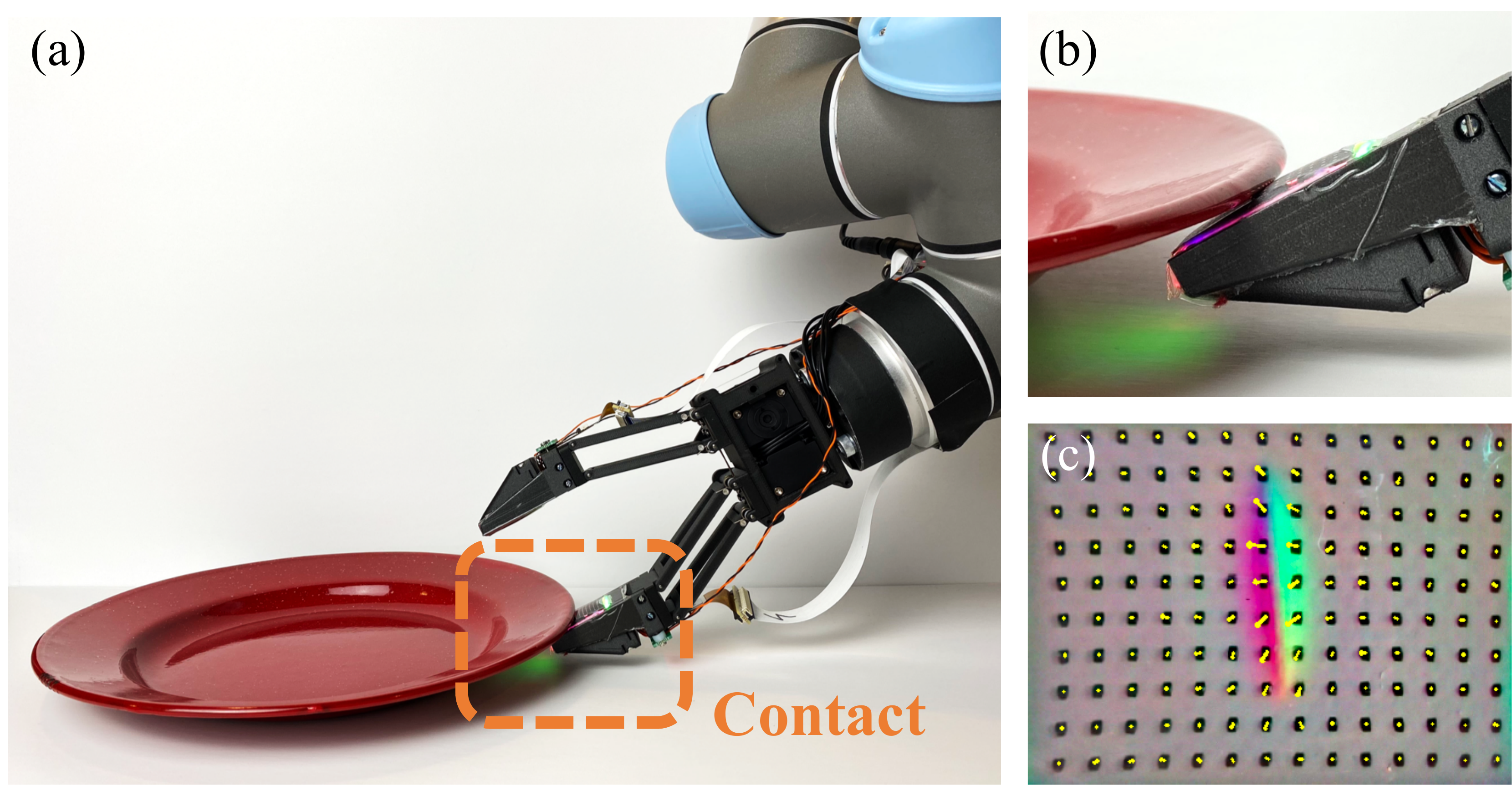}
	\caption{(a) A gripper equipped with two GelSight Wedge sensors making contact with a plate on the table; (b) A close-up view of the bottom sensor squeezing through the gap between the plate and the table; (c) The tactile imprint provides high-resolution contact geometry and forces.}
	\label{fig:teaser}
\end{figure}

In this work, we present the GelSight Wedge sensor, shown in \fig{fig:teaser}, which meets the mechanical requirements for a compact robot finger, while maintaining high-resolution 3D reconstruction. 
We introduce the related works on GelSight sensors and corresponding 3D reconstruction algorithms in Section~\ref{sec:related}. 
We then describe the design details of the sensor in Section~\ref{sec:design}. 
We evaluate 3D reconstruction with different light configurations for future design reference in Section~\ref{sec:3D}. 
Finally, we combine the reconstructed 3D point cloud with Iterative Closest Point (ICP)~\cite{besl1992method} to demonstrate its effectiveness and potential for pose tracking in the 3D space.

\section{Related Work}
\label{sec:related}

GelSight sensors are one type of vision-based tactile sensors~\cite{ward2018tactip, alspach2019soft, sferrazza2020learning, yamaguchi2016combining, lambeta2020digit, padmanabha2020omnitact, romero2020soft}. Typical vision-based tactile sensors obtain high-resolution contact information by watching the deformation of soft gels with cameras when in touch. Contact information includes forces and geometry. Tracking the motion of visual features on the gel provides contact forces ~\cite{yuan2017gelsight, ward2018tactip, yamaguchi2016combining}. In addition, GelSight sensors use reflective silicone paints and directional lights to get high-resolution 3D contact geometry applying photometric stereo~\cite{yuan2017gelsight}.

Robots can use contact geometry and forces for various robotic tasks. Contact geometry has been applied for pose estimation~\cite{li2014localization, bauza2019tactile}, material perception~\cite{yuan2018active}, grasp adjustment~\cite{calandra2018more, hogan2018tactile}, contour following~\cite{lepora2019pixels, hellman2017functional}, and ball manipulation~\cite{tian2019manipulation, lambeta2020digit}. Researchers also combine contact geometry with forces for slip detection~\cite{dong2019maintaining}, hardness measurement~\cite{yuan2017shape}, cube manipulation~\cite{hogan2020tactile}, and cable manipulation~\cite{she2019cable}.


In this work, we follow the path of GelSlim~\cite{donlon2018gelslim}, to make a compact robot finger, but keep measuring high-resolution 3D contact geometry, to provide more accurate information for robotic applications.


For 3D reconstruction algorithms, the previous GelSight sensor~\cite{dong2017improved} uses the photometric stereo technique. It maps the color to surface gradients with a lookup table, and integrates the gradients with a fast Poisson solver~\cite{fastPoisson} to get the depth. The mapping requires distinguishable color signals for different surface gradients, and the fast Poisson solver requires surface gradients in both horizontal and vertical directions. The current GelSlim design can not reach the two requirements, so the authors apply a neural network for 3D reconstruction~\cite{bauza2019tactile}. Since it is a challenging task to reconstruct 3D from raw images, it requires thousands of calibration images to learn a model that can generalize to different shapes. In this work, we design the sensor with care to meet the two requirements for photometric stereo. We demonstrate that it achieves higher-fidelity 3D reconstruction with only 30 calibration images.


\section{Design and Fabrication}
\label{sec:design}

\begin{figure}[t]
	\centering
	\includegraphics[width= \linewidth]{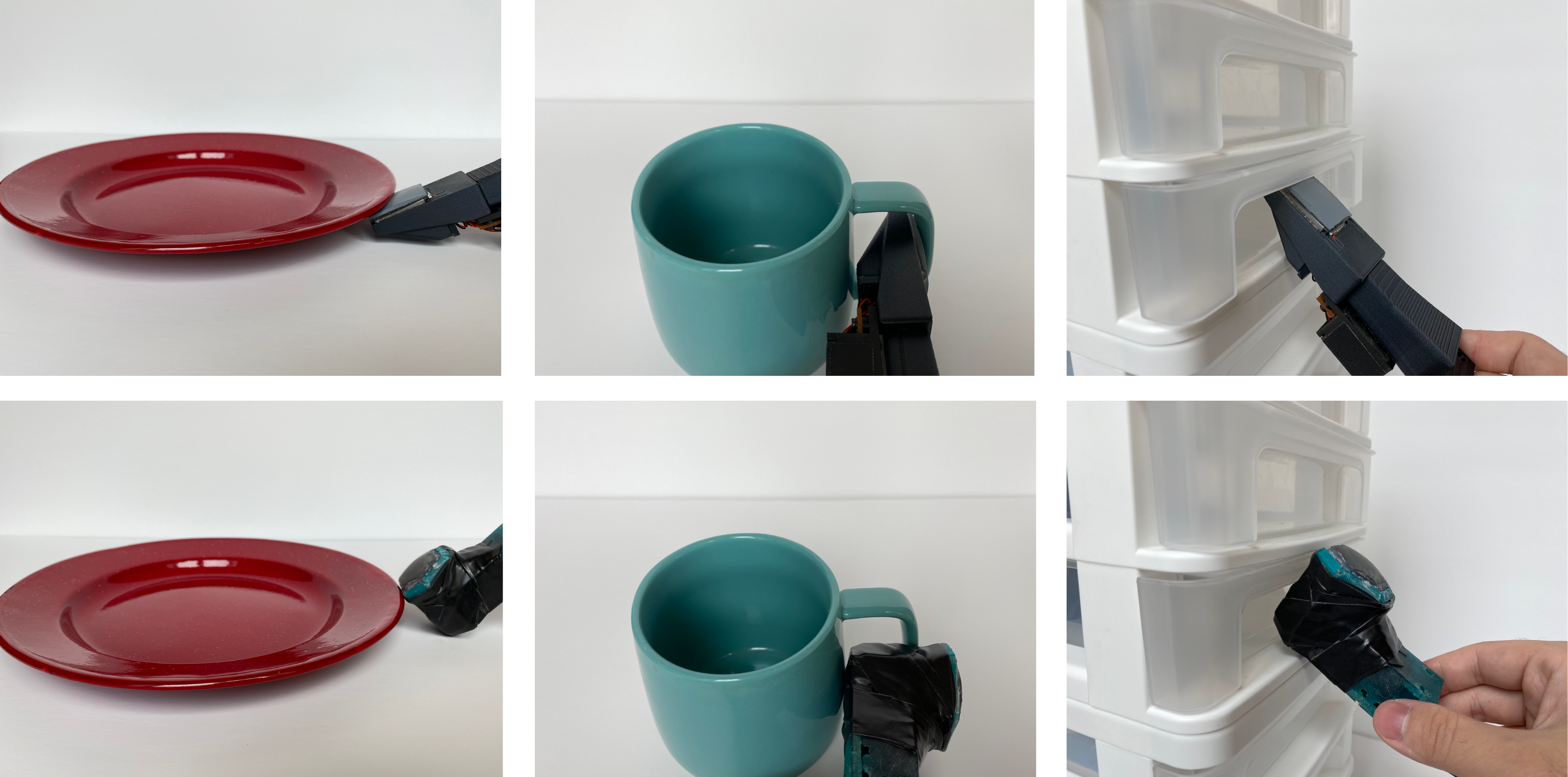}
	\caption{The compact Wedge sensor (top) can squeeze into narrow spaces for robotic tasks, compared with the previous GelSight sensor~\cite{dong2017improved} (bottom). 
	}
	\label{fig:demo}
\end{figure}

\begin{figure}[t]
	\centering
	\includegraphics[width= \linewidth]{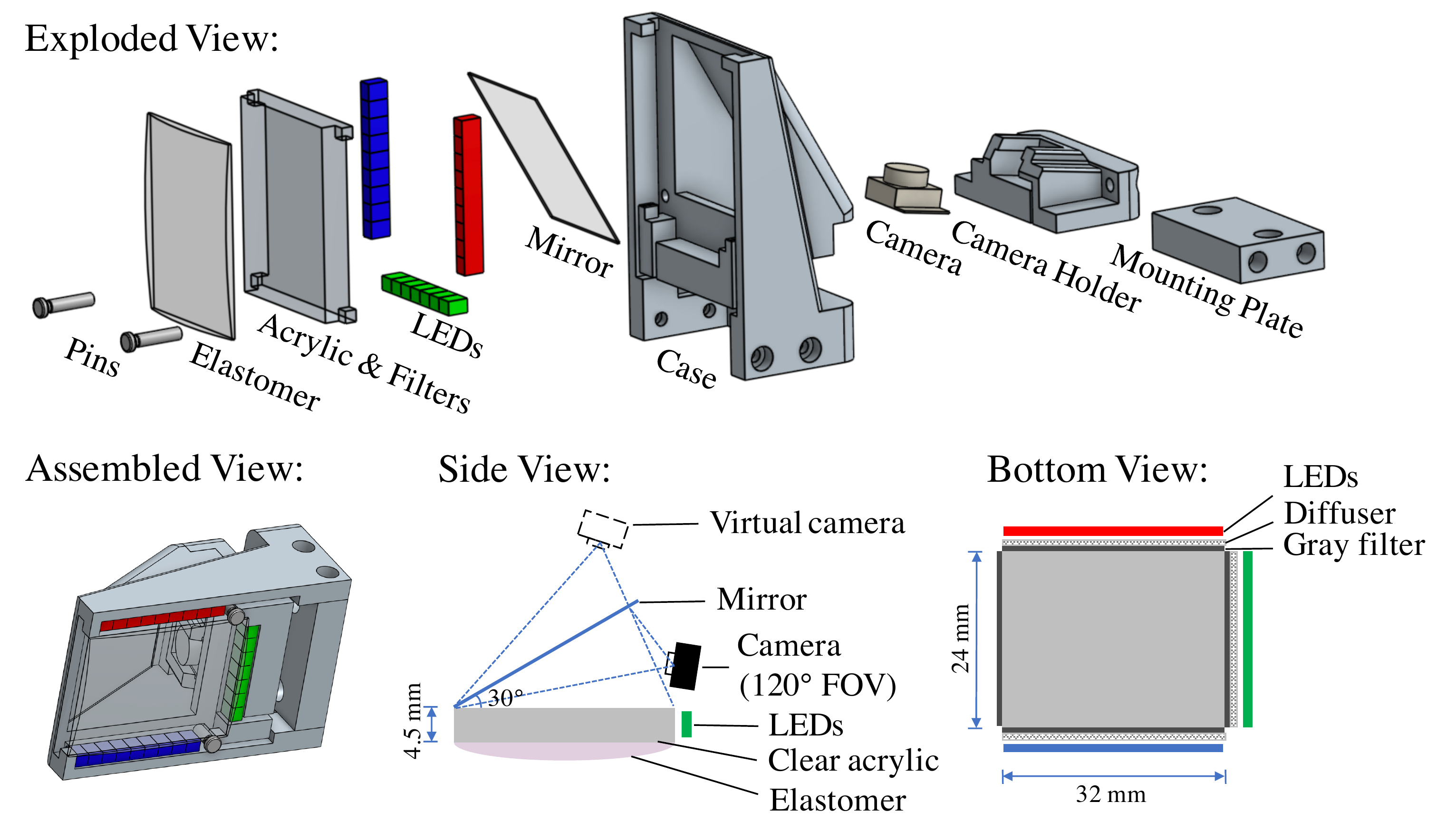}
	\caption{Sensor design, including the exploded view showing inner components, the assembled CAD model, and the schematics.}
	\label{fig:schema}
\end{figure}

Our design goal is to have a compact shape for robotic tasks, as shown in \fig{fig:demo}, while maintaining high-resolution 3D reconstruction.
To achieve this, we comprise some extent of optical properties for better shape. 
\fig{fig:schema} shows components and schematics of the sensor design. Following, we describe the design principles and learned lessons for each sensor component.
%

\myparagraph{Lighting}
The ideal lighting for GelSight 3D reconstruction would be 3 directional color lights evenly spread around 360 degrees, as \cite{dong2017improved, johnson2011microgeometry}. It can measure surface gradients from all directions. 
%

%
We sacrifice the light at the tip to pick up small objects, which discard half of the horizontal gradients. To compensate that, we put lensless LED arrays directly on the 3 sides of the acrylic to generate directional lights while keep the sensor thin, and apply filters to estimate the gradients from other 3 directions as accurate as possible.
We will demonstrate that a combination of these components can still reconstruct 3D with high-fidelity in Section~\ref{sec:3D}.
%
%
We use lensless LED with brand ``Chanzon 3528'' (3.5$\times$2.8 mm) for our sensor.

\myparagraph{Filters}
\begin{figure}[t]
	\centering
	\includegraphics[width= \linewidth]{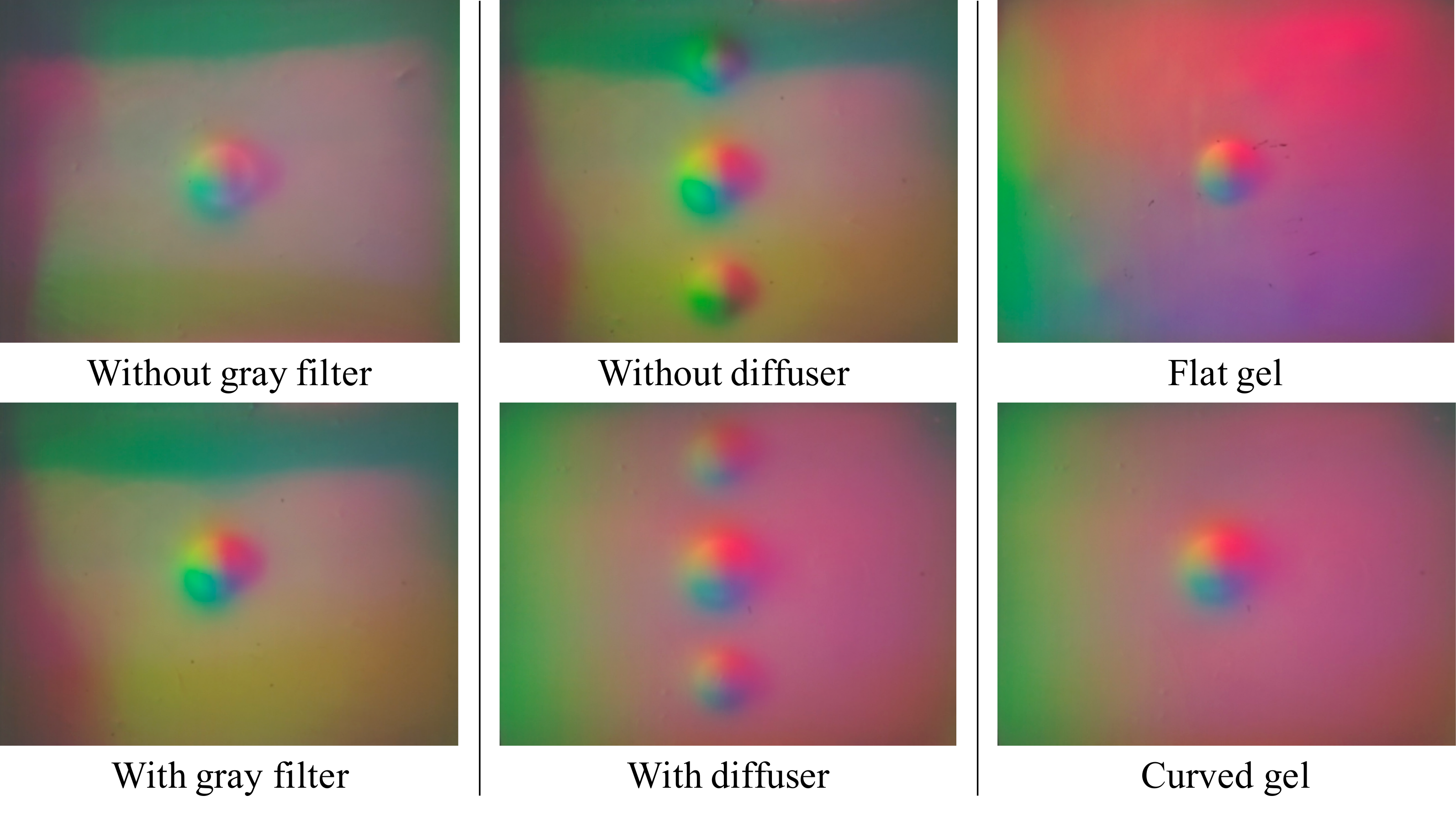}
	\caption{The effects of gray filters, diffusers, and curved gels. Gray filters increase the color contrast to better map to the surface gradients; Diffusers remove shadows and enable color gradients on the boundary (e.g. without diffusers, there is no blue lights for the bottom sphere); Curved gels produce more uniform global illumination.}
	\label{fig:filters}
	\vspace{-10px}
\end{figure}
We find gray filters and diffusers can effectively increase the contrast and uniformity of tactile imprints. We stick the filters to the side of the acrylic block, as shown in \fig{fig:schema}, bottom view. The effects are shown in \fig{fig:filters}. 

The gray filter needs to be optically coupled to the acrylic block so that there is no reflection from the air interface. One easy way is to use automobile window tint film, which comes with a layer of clear adhesive on one side. We attach the filter to the side of the acrylic block and press it carefully in place so there are no air bubbles. We use a brand called ``VViViD Air-Tint Dark Air-Release Vinyl Wrap Film'' for the sensor.

Gray filters can reduce internal reflections. When lights reflect inside the acrylic, it decreases the contrast of contact regions and the sensitivity for mapping from RGB color to the surface gradient. Gray filters can alleviate the problem. A gray filter absorbs light, which reduces the illumination level and also reduces internal reflections. The internal reflected light has to pass through the filter twice before returning, while the direct light from the LED's has to pass through it only once. Thus if the gray filter passes 1/4 of the light, then the internal reflections will be reduced to 1/16. 

Diffusers can lead to more uniform global illumination. Without diffusers, regions on the boundary are not illuminated, as shown in \fig{fig:filters}. Contact surfaces lose part of the gradient information inside these regions, causing errors in 3D reconstruction. Diffusers help to spread the light smoothly over the whole sensing area. We use ``3M Diffuser 3635-70'' for the sensor.

\myparagraph{Camera}
We use a Raspberry Pi mini camera, with a high FOV of 120 degrees, to make the sensor compact. The camera only costs \$16. We stream the video by mjpg\_streamer on the Raspberry Pi, and parse the mjpeg video stream from python on the computer. It works at 60 FPS for 640$\times$480 pixels, and 90 FPS for 320$\times$240 pixels. The default mjpg\_streamer setting can cause flickering images. To get stable and well-blend color images, it is important to turn off the AWB (auto white balance), and tune fixed gains for awbgainR (red channel), awbgainB (blue channel), and EV (exposure compensation for RGB channels).

\myparagraph{Elastomer}
The clear silicone gel can be coated with gray (Lambertian) or silver (Semi-specular) paints. In this work, we focus on 3D reconstruction with Lambertian for it provides more accurate gradient information, as suggested in \cite{dong2017improved}. We use ``Silicone Inc. XP-565'' for clear silicone, and "Print-On Silicone Ink" for paints. 

The shape of the silicone gel can be flat or curved. Optically, the curved gel provides more uniform illumination, as shown in \fig{fig:filters}. Because the curve make the lights shoot from tilted angles to more direct angles as distance grows. This compensates for the decreased LED lights over distance. Mechanically, flat gel can pick up smaller objects at the tip, while curved gel is more tolerant to flat and concave surfaces since it does not require the sensed surface to be perfectly aligned to the sensor surface. In our experiment, the shape of curved gel is a part of an ellipsoid (a=b=60mm, c=100mm).



\myparagraph{Covering}
\begin{figure}[t]
	\centering
	\includegraphics[width= \linewidth]{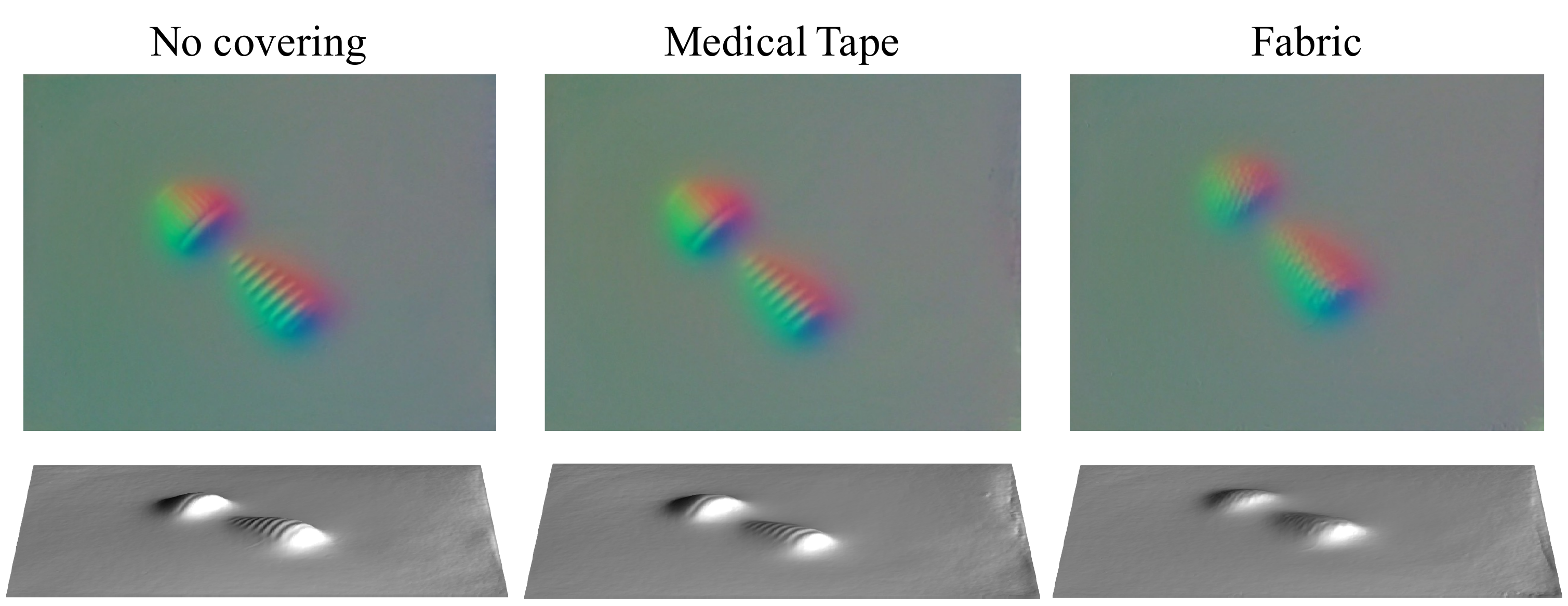}
	\caption{Comparison of different covering. The medical tape reserves more details, has lower friction forces, and medium-level of protection. It is more suitable for sliding motions for inspection and light-weight manipulation tasks. The fabric has higher friction forces, and is more rugged, but loses some extent of sensitivity. It is more suitable for heavy-weight grasping and manipulation tasks that do not require fine details. It would be ideal to switch "gloves" automatically for different tasks in the future.}
	\label{fig:covering}
\end{figure}
To improve the durability and reduce friction, the gel can be covered with ``gloves''. Besides the fabric in GelSlim~\cite{donlon2018gelslim}, we also find medical tapes helpful. Medical tapes are smooth and sensitive. Compared to fabric, the medical tape reserves much more contact details, as shown in \fig{fig:covering}. The medical tape has low friction coefficient, making it not suitable for heavy lifting application, but suitable for sliding motions for inspection or manipulation tasks. It provides medium-level protection compared to fabric. 
%
We use ``3M Tagaderm'' medical tapes for our experiments.

\myparagraph{Assembly}
For preparation, we 3D printed the sensor case, camera holder, and mold (for curved gel) with PLA materials. We polish the mold by sand papers to be smooth. We then laser cut the clear acrylic block, mirror, filters, and diffusers. 

The silicone gel is attached to the acrylic with Silpoxy or A-564 Medical Silicone Adhesive. Gray filters and diffusers are attached to the side of the acrylic. The acrylic can be either press-fit and fixed by pins or glued into the sensor case. 

The camera is press-fit into the camera holder. The camera holder is attached to the sensor case with M2 screws. The LEDs are soldered in arrays and taped into the sides of the sensor case. The front-surface mirror is glued inside the case, which is tilted
by 30 degrees.

To improve robustness, we follow GelSlim\cite{donlon2018gelslim}, and use a CSI-to-HDMI connector, which converts the fragile CSI ribbon cables to HDMI cables. The 3.3V from the connector can be used to supply the LEDs. We use an external PCB to connect resistors to the LEDs, to save space.



\section{Measuring 3D Geometry}
\label{sec:3D}

\begin{figure}[t]
	\centering
	\includegraphics[width= \linewidth]{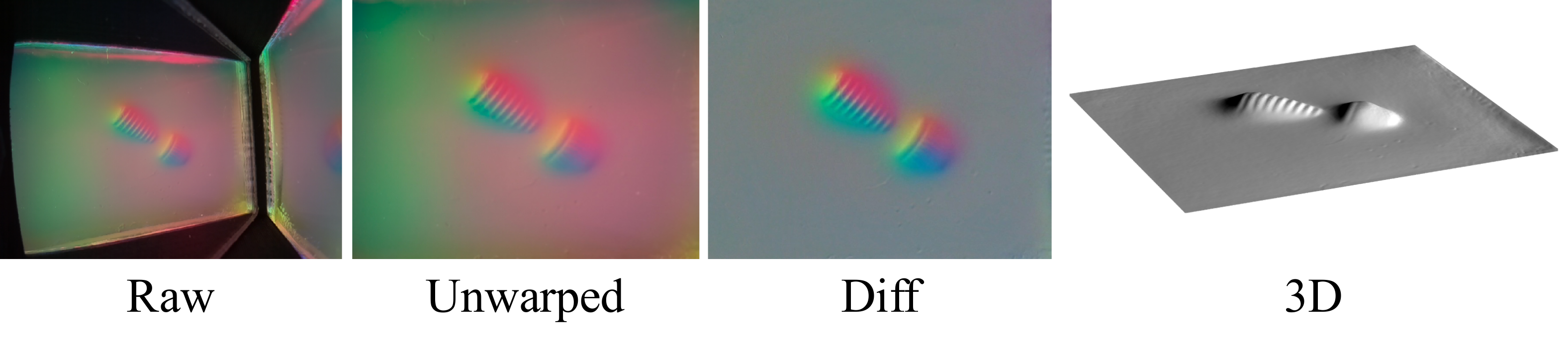}
	\caption{Workflow of 3D reconstruction. We first unwarp the raw imprint to rectangular shape; then subtract it with the blank background frame to get the difference image; finally calculate the 3D shape by mapping the difference image to surface gradients and applying fast Poisson solver for integration.}
	\label{fig:3d}
\end{figure}

\begin{figure*}[t]
	\centering
	\includegraphics[width= \linewidth]{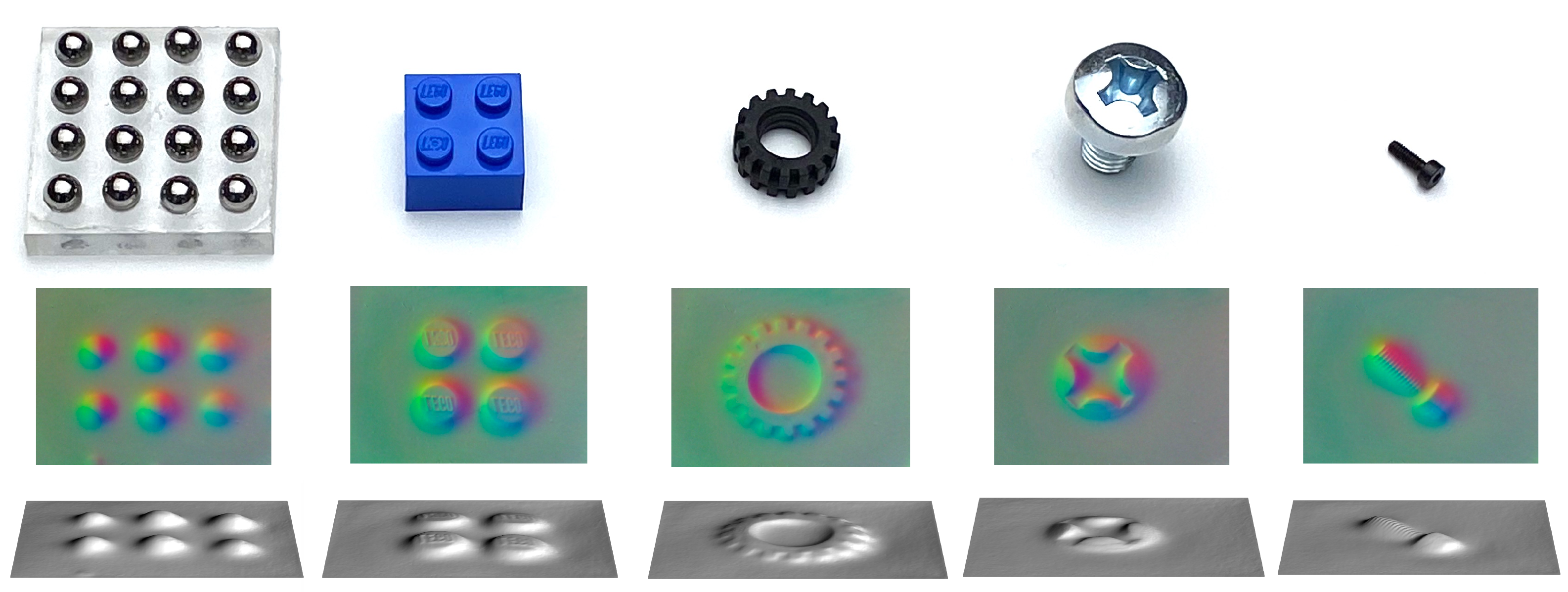}
	\caption{From top row to bottom: visual images, GelSight imprints, and inferred depth of a ball array, a Lego block, a rubber tyre, a screw cap, a M2 screw.}
	\label{fig:showcases}
	\vspace{-10pt}
\end{figure*}

In this section, we discuss the performance of 3D reconstruction with GelSight Wedge sensor. We first describe the approach for photometric stereo, and demonstrate the results with 3 lights in Section~\ref{sub:3_lights}. We then extend the method to designs with 1 or 2 lights, and compare the results with different light configurations in Section~\ref{sub:2_lights}. We finally show the effect of disturbance in Section~\ref{sub:disturbance}, and potential capabilities in Section~\ref{sub:potential}.

\subsection{3D reconstruction with 3 lights}
\label{sub:3_lights}

As described in ~\cite{yuan2017gelsight}, GelSights 3D reconstruction consists of two steps: 1) calculating the mapping from color (RGB) to horizontal and vertical surface gradients (Gx, Gy); 2) applying fast Poisson solver ~\cite{fastPoisson} to integrate gradients and get the depth.

\myparagraph{Color-to-gradient mapping}
To get the data for color-to-gradient mapping, we 1) press a calibration ball, with a known radius (2.4 mm in our case), on the sensor; 2) calculate the center and the radius of the contact circle in pixels, with Hough Circle Transform and human adjustment; 3) press a caliper on the sensor to calculate the number of pixels per millimeter. With these parameters, we can calculate the gradient Gx, Gy for each pixel in contact, and gather the pixel’s color and position: R, G, B, X, Y. Because the LED diminishes when travelling, we add the pixel location, X and Y, to compensate the light attenuation across the sensor.

To learn the mapping from color (RGBXY) to gradient (Gx, Gy), we simplify the lookup table ~\cite{dong2017improved} to a small neural network, a multi-layer perceptron (MLP) with 3 hidden layers (5-32-32-32-2), with tanh activation. Previous lookup table requires discretization to optimize efficiency and it is difficult to incorporate position. In contrast, a small neural network produces continuous results and can consider color changes across the sensor. 

We collect 32 images for training, and 8 images for testing. Besides the pixels in contact, we add 5\% of pixels not in contact (whose gradients should be zeros) to balance the data. 

\myparagraph{3D reconstruction with fast Poisson solver}
After getting the gradients Gx, Gy, we use a 2D fast Poisson solver to calculate the depth. We use a python version\cite{fastPoisson}. It takes in Gx, Gy, and boundary conditions and returns the depth. 
%
We set the boundaries to zeros for best approximation. 

\fig{fig:showcases} shows some qualitative results of reconstructed 3D of different objects: a ball array, a Lego block, a rubber tyre, a screw cap, and a M2 screw. Compared to previous work on GelSlim~\cite{bauza2019tactile}, we can reconstruct the surface with more fidelity. 
%
%
Following, we analyze and compare the 3D reconstruction results with different light configurations.

\subsection{3D reconstruction with 1 or 2 lights}
\label{sub:2_lights}
It can be difficult to place all 3 lights for some devices, e.g. GelFlex~\cite{she2019exoskeleton}, where one axis of the sensor is too long and deformable for directional lights to travel. We aim to still estimate 3D geometries with the limited lights.

\begin{figure}[t]
	\centering
	\includegraphics[width= \linewidth]{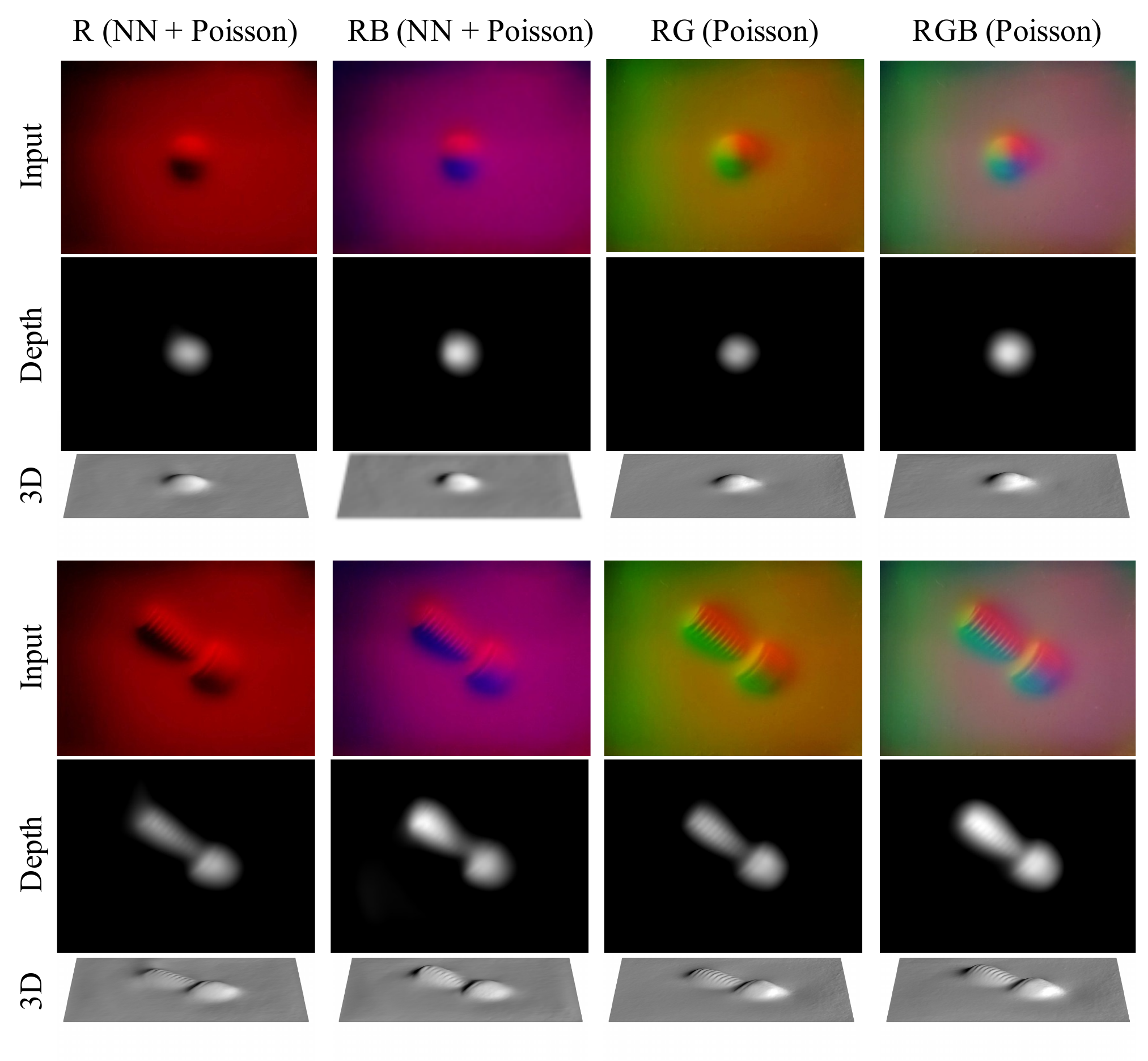}
	\caption{3D reconstruction with different light configurations. With the perpendicular lights (e.g., RGB and RG), Poisson alone can 3D reconstruct with high fidelity. Without the perpendicular lights (e.g., RB and R), neural networks can be applied to estimate the horizontal gradients and improve the reconstruction results.}
	\label{fig:light_comparison}
	\vspace{-5px}
\end{figure}

\myparagraph{Perpendicular lights}
%
%
The perpendicular lights, such as RG, provide gradients for both axes, allowing us to keep using fast Poisson solver. 
Although it misses half of gradients for each of horizontal and vertical axes, the fast Poisson solver can still integrate the gradients reasonably. It generates comparable results to RGB lights qualitatively as shown in \fig{fig:light_comparison}, and quantitatively as shown in \tab{tab:gradient}.

However, if we miss the light completely from one axis, it becomes more challenging to 3D reconstruct. Following, we will describe the challenges and possible solutions.

\myparagraph{Challenge }
The fast poisson solver requires the gradient of both directions, Gx, and Gy. When missing lights in one axis, we lose the corresponding gradient information. Without losing generality, assume we lose the green lights for our sensor, which corresponds to horizontal gradients Gx.

\begin{figure}[ht]
	\centering
	\includegraphics[width= \linewidth]{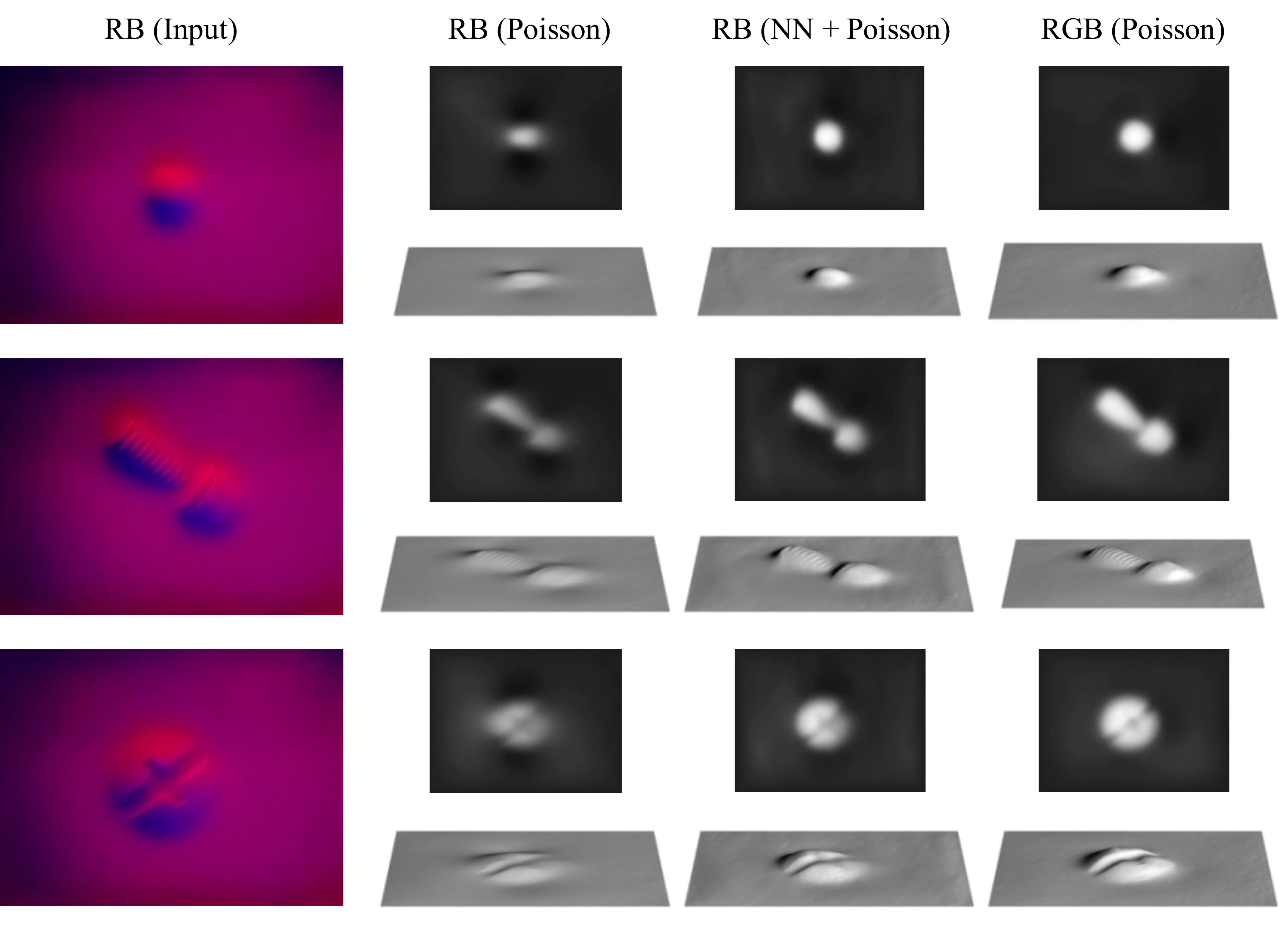}
	\caption{3D reconstruction with two lights from opposite directions (RB). Without the horizontal gradients, fast Poisson solver reconstruct depth with horizontal blur and negative value. With NN-estimated horizontal gradients, it significantly improve the 3D reconstruction results, which are close to the results of RGB lights, without extra real data. Depth images are added with a constant value in this figure to show the negative depth for RB (Poisson).}
	\vspace{-5px}
	\label{fig:twolights}
\end{figure}

A naive approximation would be assuming the gradient Gx is all zeros, and sending it to fast Poisson solver. As shown in \fig{fig:twolights}, this causes the reconstructed 3D to be “flattened” in the x-axis, and generates negative depth to meet the constraints.

Alternatively, we can leverage neural networks (NN). One way would be directly mapping from raw images to depth, as described in ~\cite{bauza2019tactile}. However, 3D reconstruction from raw images is complex, which involves both learning the color-to-gradient mapping, and solving partial derivative equations. As a result, it requires many tactile imprints pressing on various shapes to generalize to different shapes and sensors.

We hope to find a more data-efficient method that uses only a handful of calibration data. We approach this by combining NN with fast Poisson solver and Sim2real techniques.

\begin{table*}[ht]
\centering
\caption{Gradient estimation errors with different light configurations}
\begin{tabular}{|c|c|c|c|c|c|c|}
\hline
                & R (w/o NN) & R (w/ NN) & RB (w/o NN) & RB (w/ NN) & RG (w/o NN) & RGB (w/o NN) \\ \hline
$Gx$ error        & 0.104       & 0.055            & 0.106        & 0.051             & 0.045        & 0.043         \\ \hline
$Gy$ error        & 0.055       & 0.055            & 0.045        & 0.045             & 0.044        & 0.042         \\ \hline
$\theta x$ error (degs) & 5.598       & 2.899            & 5.706        & 2.656             & 2.331        & 2.248         \\ \hline
$\theta y$ error (degs) & 2.889       & 2.897            & 2.349        & 2.351             & 2.293        & 2.187         \\ \hline
\end{tabular}
\label{tab:gradient}
\end{table*}

\myparagraph{Gradient estimation for the missing axis}
The fast Poisson solver generates inaccurate results when receiving horizontal gradients Gx as zeros, as shown in \fig{fig:twolights}. But we have the vertical gradients Gy. We can improve Gx estimation with the information of Gy.

Inferring Gx from Gy is a generic problem independent of real or synthesized data. Therefore, we can use synthesized data to reduce the required real data. We synthesized 10k random depth images, and calculated gradients (Gy as input, Gx as output) for training a neural network model. We then send the inferred Gx along with Gy to Fast Poisson Solver. As shown in \fig{fig:twolights}, it significantly improves the 3D reconstruction with the inferred Gx, without any additional real data.

We use U-Net~\cite{ronneberger2015u} as our neural network model, which has been widely used for semantic segmentation. U-Net follows the standard encoder-decoder framework, and adds multi-scale skip passes to provide more details.

For the synthesis process, we generate random surfaces to learn a generic model. We flatten the boundaries and remove negative depth to match the real data. For details, we generate random arrays ranging from -1 to 1 and cube the values for magnification; set boundaries to zeros; apply Gaussian filters with random kernel sizes; clip the negative depth; and scale to random maximal height within the range of real sensors. We find it important to set boundaries to zeros to match the real sensors and provide constraints. Otherwise, it can generate artifacts like small bumps between contact regions to the boundary.


\myparagraph{Results}
We compare the 3D reconstruction results with different methods and different light configurations. 

\fig{fig:twolights} shows the effectiveness of applying neural networks for estimating the gradients Gx. Without neural networks, the fast Poisson solver can only take Gx as zeros, and generates depth with horizontal blur and negative values. With neural networks to estimate missing gradients, it significantly improves the 3D reconstruction, without any extra real data.

\fig{fig:light_comparison} shows the qualitative comparison of 3D reconstruction with different light configurations. We find that the RG can perform surprisingly well with the fast Poisson solver alone. The half gradients of Gx and Gy already provide many constraints for the fast Poisson solver. In contrast, the RB requires neural networks to estimate the Gx from Gy to achieve comparable results. It is more challenging to 3D reconstruct with only 1 light, it tends to generate artifacts like small bumps between the contact region to the boundary due to too many missing gradients. We hope to refine the results for 1 light with a small amount of real data in the future. 

\tab{tab:gradient} shows the quantitative results for gradient estimation with different light configurations. We use the collected 40 real images with a calibration ball, and split 32 for training, 8 for testing. The UNet uses 10k synthesized data with the described process. The results agree with the qualitative evaluation. In summary, the UNet can significantly improve the gradient estimation for the missing axis without extra real data. The preference order for light configurations is: RGB (3 lights) $>$ RG (2 perpendicular lights) $>$ RB (2 opposing lights) $>$ R (1 light).

\subsection{Influence of disturbances}
\label{sub:disturbance}
\begin{figure}[t]
	\centering
	\includegraphics[width=0.9 \linewidth]{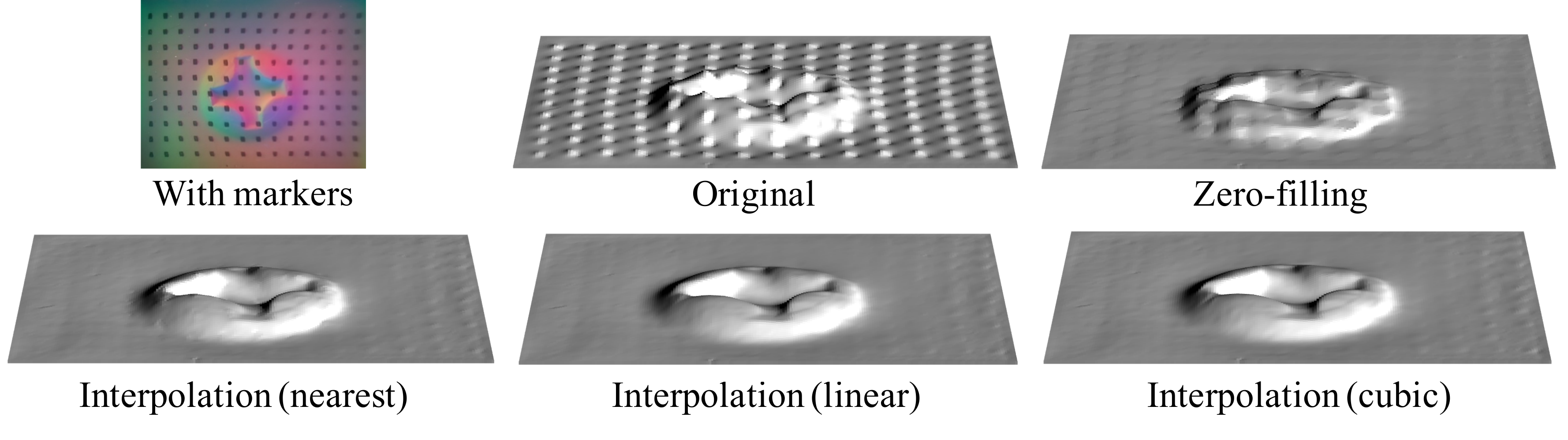}
	\caption{Comparison of interpolation methods for 3D reconstruction with markers.}
	\label{fig:markers}
	\vspace{-10pt}
\end{figure}

\myparagraph{Markers} We can add markers on the gel to observe force distributions. However, the marker will influence the 3D reconstruction as shown in \fig{fig:markers}. Without extra processing, the markers result in small bumps in 3D. We can alleviate it by filling zeros gradients to the markers. It removes bumps and runs fast, but generates flat patches. We can improve the result by interpolating the gradients for the markers. In our experiments, we use $griddata$ in the scipy package~\cite{2020SciPy-NMeth} for interpolation. To run in real time, we interpolate with the ‘nearest’ method, and only sample the pixels around the markers. 
It generates reasonable interpolation results and only take 10 ms (for resolution of 200x150), allowing real-time feedback control. In contrast, 'linear'/'cubic' generate better results but take 60/70 ms.

\begin{figure}[t]
	\centering
	\includegraphics[width=0.9 \linewidth]{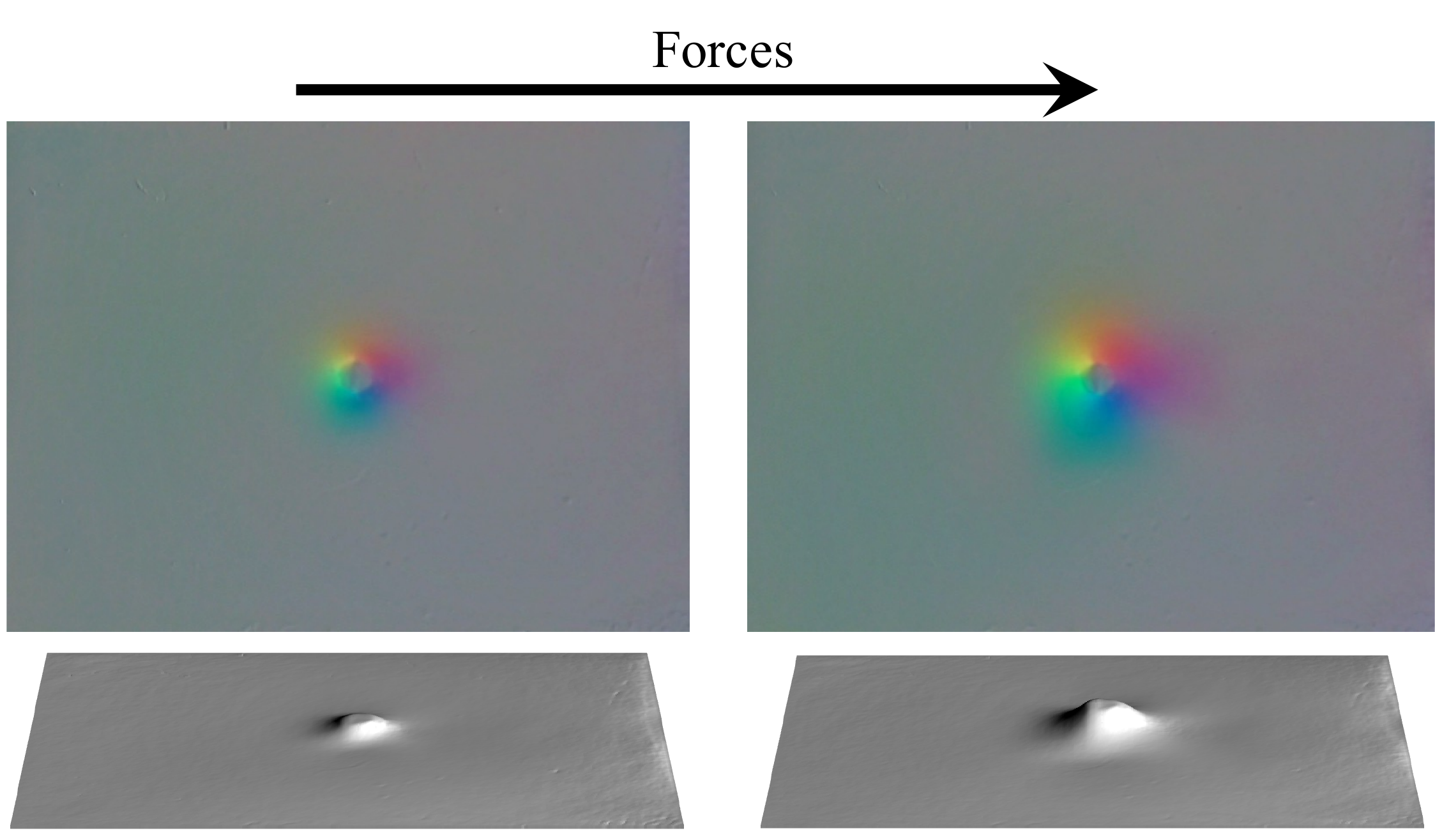}
	\caption{The gel deformation and shadows can influence the 3D reconstruction. We press a screw against the sensor with different forces. We can observe that the reconstructed shape does not exactly follow the object shape. It requires further processing from the perception side in the future.}
	\label{fig:shadows}
	\vspace{-10pt}
\end{figure}

\myparagraph{Gel deformation and shadows} \fig{fig:shadows} shows the gel deformation and shadows can influence the 3D reconstruction, especially for sharp surfaces. It bring challenges to extract the surface of the object from reconstructed 3D shape, but also make the signals force-dependent.


\subsection{Potential capabilities}
\label{sub:potential}

\begin{figure}[t]
	\centering
	\includegraphics[width=\linewidth]{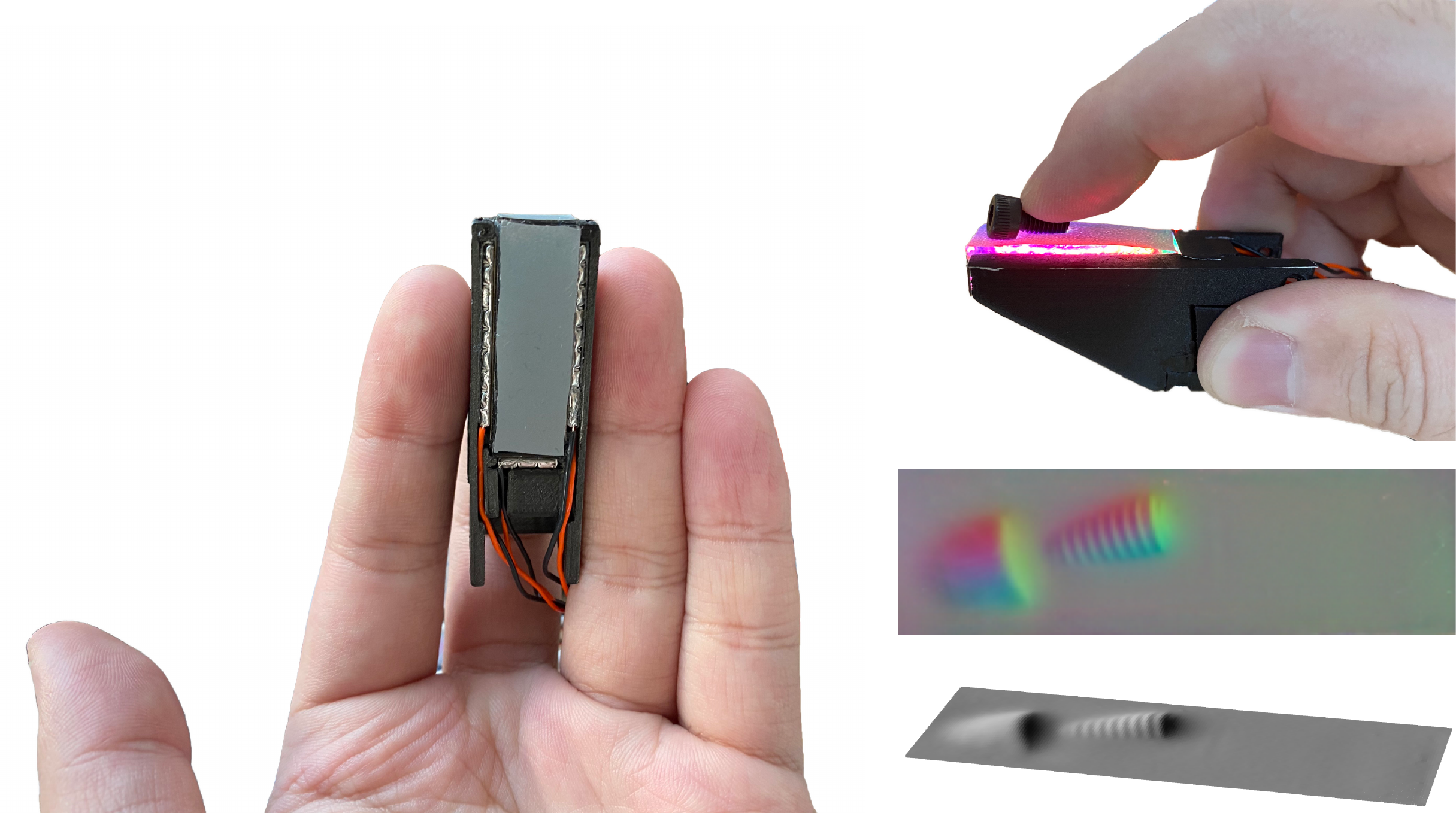}
	\caption{Shrinking the sensor to the size of human fingers. Left shows the comparison with human fingers. Right shows the visual image, the GelSight difference image, and reconstructed 3D of touching a screw. The finger-size sensor can be applied for fine manipulation tasks, and multi-finger hands.}
	\label{fig:finger}
\end{figure}

\myparagraph{Flexibility} We demonstrate the design's flexibility by shrinking it to the size of a human finger, while retaining high-resolution 3D reconstruction, as shown in \fig{fig:finger}. It is more suitable for fine manipulation tasks, and multi-finger hands in the future.

\begin{figure}[t]
	\centering
	\includegraphics[width= \linewidth]{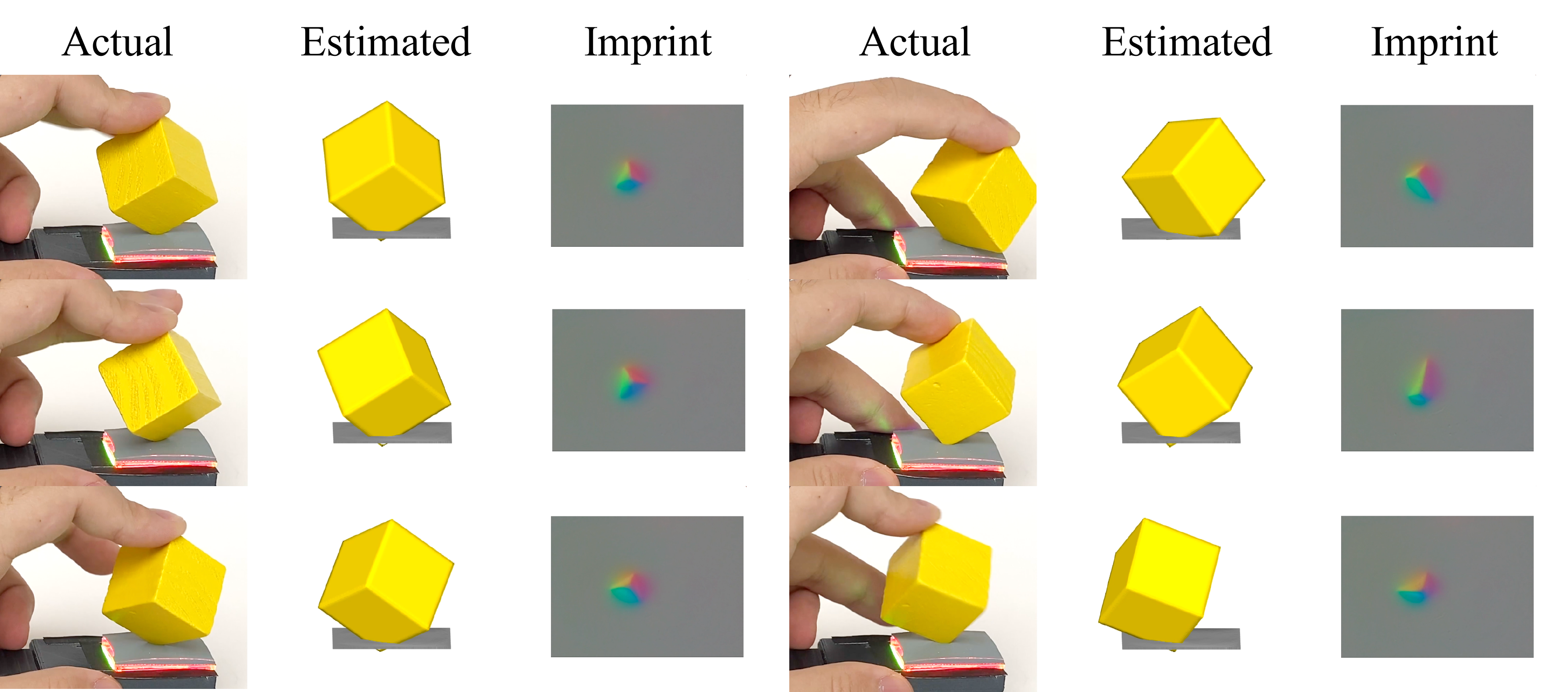}
	\caption{Tracking the pose of a cube by corners with ICP and reconstructed 3D point clouds from touch alone.}
	\label{fig:pose}
\end{figure}

\myparagraph{Pose tracking} The reconstructed 3D point cloud makes it convenient to apply ICP for pose tracking~\cite{izatt2017tracking, alspach2019soft}. We demonstrate its effectiveness and potential by an example of tracking the pose of a cube by its corner. \fig{fig:pose} shows that it accurately tracks the poses with various orientations. We assume the initial pose is given. We use the ICP algorithm from Open3D~\cite{Zhou2018}. It runs at about 10 Hz. Detailed analysis of pose error and complex geometry is out of the scope of this paper.

\section{Conclusion}
\label{sec:conclusion}

In this work, we present the GelSight Wedge sensor, which can measure high-resolution 3D contact geometry in a compact shape. We illustrate the effect of the components in our design, and evaluate the 3D reconstruction with different light configurations. For applications, we successfully shrink the sensor to the size of a human finger to work for narrow environments and fine manipulation tasks in the future. We also demonstrate the effectiveness and potential of the reconstructed 3D point cloud for pose tracking.

In the future, we hope to further reduce the influence of markers on 3D reconstruction. It can be achieved by reducing the size of markers, or changing black dots to checkerboard of lighter and darker gray. The acrylic currently is 4.5mm thick, which is decided by the LEDs size. Replacing LED arrays with smaller sizes can further reduce the thickness of the finger tip. We hope to combine the sensor with robotic grippers and multi-finger hands towards more reactive and robust control for various grasping and manipulation tasks in the future.



\section*{ACKNOWLEDGEMENTS}

Toyota Research Institute (TRI), and the Office of Naval Research (ONR) [N00014-18-1-2815] provided funds to support this work. The authors would also like to thank Achu Wilson, Alex Alspach for the help during prototyping, and Alberto Rodriguez, Siyuan Dong, Neha Sunil for insightful discussions.

\bibliographystyle{IEEEtran}
\bibliography{ref}

\end{document}